\documentclass{article}

\usepackage[preprint]{neurips_2022}

\usepackage[utf8]{inputenc} 
\usepackage[T1]{fontenc}    
\usepackage{hyperref}       
\usepackage{url}            
\usepackage{booktabs}       
\usepackage{amsfonts}       
\usepackage{nicefrac}       
\usepackage{microtype}      
\usepackage{xcolor}         
\usepackage{amsmath}
\usepackage{bm}
\usepackage{graphicx}
\usepackage{subfigure}
\usepackage{multirow}
\newtheorem{definition}{Definition}
\newtheorem{theorem}{Theorem}
\newtheorem{assumption}{Assumption}

\usepackage{algorithm2e}
\usepackage{wrapfig}

\title{Sequence-to-Set Generative Models}

\author{%
  Longtao Tang\\
  School of Data Science\\
  City University of Hong Kong\\
  Kowloon, Hong Kong \\
  \texttt{longttang2-c@my.cityu.edu.hk} \\
  \And
  Ying Zhou \\
  Department of Economics and Finance\\
  City University of Hong Kong\\
  Kowloon, Hong Kong \\
  \texttt{ying.zhou@cityu.edu.hk} \\
  \And
  Yu Yang\thanks{Corresponding author} \\
  School of Data Science\\
  City University of Hong Kong\\
  Kowloon, Hong Kong \\
  \texttt{yuyang@cityu.edu.hk} \\
}

\begin{document}

\maketitle

\begin{abstract}
In this paper, we propose a sequence-to-set method that can transform any sequence generative model based on maximum likelihood to a set generative model where we can evaluate the utility/probability of any set. An efficient importance sampling algorithm is devised to tackle the computational challenge of learning our sequence-to-set model. We present GRU2Set, which is an instance of our sequence-to-set method and employs the famous GRU model as the sequence generative model.
To further obtain permutation invariant representation of sets, we devise the SetNN model which is also an instance of the sequence-to-set model. A direct application of our models is to learn an order/set distribution from a collection of e-commerce orders, which is an essential step in many important operational decisions such as inventory arrangement for fast delivery. Based on the intuition that small-sized sets are usually easier to learn than large sets, we propose a size-bias trick that can help learn better set distributions with respect to the $\ell_1$-distance evaluation metric. Two e-commerce order datasets, TMALL and HKTVMALL, are used to conduct extensive experiments to show the effectiveness of our models. The experimental results demonstrate that our models can learn better set/order distributions from order data than the baselines. Moreover, no matter what model we use, applying the size-bias trick can always improve the quality of the set distribution learned from data.

\end{abstract}

\section{Introduction}
Learning a generative model~\citep{Goodfellow14, Kingma14} from data to generate random samples is a fundamental unsupervised learning task. In many real applications, the data, such as customer orders and drug packages, can be represented by sets. Learning a set generative model can boost many important applications such as simulating users' demand of e-commerce orders for inventory arrangement~\citep{Tongwen21}. However, in literature, little attention has been paid to generative models for sets.


In this paper, we study the problem of learning a set generative model from data. Specifically, we study sets containing some categorical items from a finite ground set. 
Some previous studies~\citep{Benson18, Tongwen21} proposed traditional statistical models to learn the set generative model, while how to employ deep learning to design set generative models remains largely untouched. Meanwhile, deep generative models for sequence data~\citep{hochreiter1997long, Maximilian06, Kyunghyun14, devlin2018bert} have been extensively studied. Since the only difference between a sequence and a set is that the order of items does not matter in a set, we propose to leverage a sequence generative model to build a set generative model. 

A natural idea of converting a random sequence to a random set is to just ignore the order of items. However, for a set, the number of possible sequences that can be regarded as equivalent to the set is exponential to the number of items. This brings us a serious computational challenge where we need to enumerate all the equivalent sequences for a set when learning the parameters of the model. Traditional statistical models~\citep{Benson18, Tongwen21} along this line either make unrealistic independent assumptions on items or run a learning algorithm of complexity exponential to set sizes to only deal with small sets. These constrained treatments undermine the model's ability to capture the complex correlations among items in sets.

To tackle the computational challenge while still well capturing the correlations among items, in this paper, we make the following technical contributions.

In Section~\ref{section:Sequence2Set}, we first show that we can convert any sequence generative model based on maximum likelihood, such as RNN~\citep{Maximilian06} and DeepDiffuse~\citep{islam2018deepdiffuse}, to a set generative model. We call our method Sequence-to-Set. To address the computational challenge, we propose an efficient importance sampling algorithm to estimate the gradient of the log-likelihood function of our Sequence-to-Set model.

In Section~\ref{sec:models}, to capture the complex item correlations, we propose two deep models GRU2Set and SetNN, which are instances of our Sequence-to-Set model.
GRU2Set directly uses the GRU as an essential building block, while SetNN is designed for obtaining permutation invariant set embeddings when learning a set generative model. We discuss that GRU2Set and SetNN can express any set distributions as long as we increase the model capacity.

To learn a better set distribution from data, in Section~\ref{section:Size Distribution}, we reveal that we often need to modify the empirical size distribution of the training data and propose a heuristic to do so. This size-bias trick can be applied to all set generative models aiming at learning a set distribution.

We conduct extensive experiments using two e-commerce datasets TMALL and HKTVMALL, and report the experimental results in Section~\ref{sec:exp}. The experimental results clearly show the superiority of our models to the baselines and the effectiveness of our size-bias trick. 

\section{Related Work}
Learning a set generative model is often regarded as learning a choice model for subset selection. Along this line, \cite{Benson18} proposed a discrete choice model (DCM) to model the utility/probability that users select a specific set. \cite{Tongwen21} utilized representation learning to design a parameterized Markov Random Walk to model how users select random sets from an item graph. Both \citep{Benson18} and \citep{Tongwen21} are set generative models, though they are not deep models. To achieve practical computational efficiency, both \citep{Benson18} and \citep{Tongwen21} only handle small sets and compromise on modeling high-order item correlations. 

To model the permutation invariant nature of sets, \cite{Francisco17} used the sum or max of item embedding as the set embedding. \cite{Zaheer17} systematically investigated the theory of permutation invariant functions. \cite{stelzner20} proposed a generative model for point clouds based on GAN~\citep{Goodfellow14} and Set Transformer~\citep{Lee19}. Although a point cloud is a set of 2D-coordinates, we cannot use the point cloud generative model~\citep{stelzner20} to solve our problem as the model outputs continuous random variables instead of categorical data. \cite{Adam20} presented another generative model for sets of real-valued vectors based on Set Transformer~\citep{Lee19} and their generating process is based on VAE~\citep{Kingma14}. A disadvantage of using GAN or VAE to train a generative model is that it would be difficult to evaluate the probability density of a random sample. Normalizing Flows are powerful generative models based on solid statistical theories. However, the study of Discrete Flows is still in its infancy \citep{Hoogeboom19, Tran19} and no existing works have investigated using Discrete Flows to learn set distributions.

Multi-label classification also needs to model the permutation invariance of sets of labels. \cite{yang2019deep} applied reinforcement learning with a permutation-invariant reward function to remove the impact of label orders. \cite{yazici2020orderless} proposed to align the labels of a sample with its predicted labels before calculating the loss of the sample. However, our task in this paper is an unsupervised learning task. We cannot apply the loss functions in~\citep{yang2019deep,yazici2020orderless} since we do not have ``labels'' (or feature vectors if one regards a set as a set of labels) for training samples. 

To sum up, existing works on set generative models either deal with sets of real-valued vectors which are different from categorical sets, or adopt non-deep models which need to compromise on capturing item correlations to make the computation efficient. It is still urgent to design deep generative models for categorical sets and propose efficient algorithms to deal with computational challenges.

\section{Sequence-to-Set Generative Models}\label{section:Sequence2Set}


In this section, we introduce how to convert a sequence generative model to a set generative model and how to optimize the model parameters.

\subsection{Converting Sequences to Sets}
We first characterize the sequence generative model that we want to convert to a set generative model.
\begin{definition}[Sequence Generative Model]\label{def:seq}
A sequence generative model is a model with a sequential generating process, which can be represented by a sequence of states and actions $[s_0, a_0, s_1, a_1, ..., s_{T-1}, a_{T-1}, s_T]$, where $s_i$ is the state at time $i$, $a_i$ is the action taken at time $i$ and $s_T$ is the end state. We call $[s_0, a_0, s_1, a_1, ..., s_{T-1}, a_{T-1}, s_T]$ a {\bf generating path}. Like a finite-state machine, the sequence generative model uses each action $a_i$ to {\bf deterministically} update the current state $s_i$ to $s_{i+1}$. For each state $s$, the sequence generative model has a corresponding probability distribution $p(a \mid s;\bm{\theta})$ indicating the probability of choosing action $a$ when the current state is $s$, where $\bm{\theta}$ is the parameters of the sequence generative model. 
\end{definition}

Our definition of the sequence generative model can cover a wide range of models that can generate items in a sequential manner. For example, the commonly used sequence model RNN is an instance, where the state is the history vector $H$ and an action is to choose the next word/item. Moreover, influence diffusion models such as the Independent Cascade (IC) model~\citep{Goldenberg01} and the DeepDiffuse model~\citep{islam2018deepdiffuse} are also covered by Definition~\ref{def:seq}, since in these models the state at time $t$ is the set of activated nodes until time $t$ and the action $a_t$ is to select the group of nodes to be activated at time $t+1$.

To convert a sequence generative model in Definition~\ref{def:seq} to a set generative model, we need to associate each state $s$ with a {\bf state-associated set} $S_s$. For example, in an RNN, we can define $S_{s_t}$ as the set of items generated in the first $t$ steps. In the IC model, we can define $S_{s_t}$ as the set of nodes activated at or before time $t$. 


We say that a generating path $l=[s_0, a_0, \dots, s_{T-1}, a_{T-1}, s_T]$ can {\bf induce} a set $S$ if $S=S_{s_T}$. Let $L(S)=\{l \mid l\text{ can induce }S\}$ be the {\bf path set} of $S$. As the sequence generative model has the probability distribution $p(a \mid s;\bm{\theta})$ indicating the probability of taking $a$ as the next action given the current state $s$, we can derive the probability of having a path $l=[s_0, a_0, \dots, s_{T-1}, a_{T-1}, s_T]$ as
\begin{equation*}
    p(l;\bm{\theta}) = p(a_0 \mid s_0;\bm{\theta}) \times p(a_1 \mid s_1;\bm{\theta}) \times \cdots \times p(a_T \mid s_T;\bm{\theta}).
\end{equation*}
Therefore, given a sequence generative model with parameters $\bm{\theta}$, the probability to induce a set $S$ is
\begin{equation*}
p(S;\bm{\theta}) = \sum_{l \in L(S)} p(l;\bm{\theta}).
\end{equation*}

\subsection{Parameter Learning}

Suppose we have a collection (multiset) of observed sets $\mathcal{S}=\{S_1,S_2,...,S_N\}$. We want to use $\mathcal{S}$ as the training data to learn a sequence model which can generate the observed sets. We adopt maximum likelihood estimation to learn the parameters $\bm{\theta}$. Specifically, the loss function is
\begin{equation*}\small
\mathcal{L}(\bm{\theta}) = - \sum_{i=1}^N \log p(S_i; \bm{\theta})=-\sum_{i=1}^N\log \left( \sum_{l \in L(S_i)}p(l;\bm{\theta}) \right)
\end{equation*}
We apply stochastic gradient descent based algorithms to minimize the loss function. The stochastic gradient w.r.t. an observation $S$ is $\nabla \log{p(S;\bm{\theta})}=\nabla \log \left( \sum_{l \in L(S)} p(l;\bm{\theta}) \right)$. As in the log-sum term of $\nabla \log p(S;\bm{\theta})$, the number of paths that can induce $S$ is often exponential to $|S|$ (in some sequence generative model the number of such paths is even infinite), it is very challenging to evaluate $\nabla \log{p(S;\bm{\theta})}$ exactly.

To tackle the computational challenge in computing $\nabla \log{p(S;\bm{\theta})}$, we first find that $\nabla \log{p(S;\bm{\theta})}$ can be regarded as an expectation as follows.
\begin{equation*}\small
  \nabla \log p(S;\bm{\theta}) = \frac{\sum_{l \in L(S)} \nabla p(l;\bm{\theta})}{p(S;\bm{\theta})} = \frac{ \sum_{l\in L(S)} p(l;\bm{\theta}) \nabla \log p(l;\bm{\theta}) }{p(S;\bm{\theta})}
=E_{l\sim p(l|S;\bm{\theta})} [\nabla \log p(l;\bm{\theta})], 
\end{equation*}
where $p(l \mid S;\bm{\theta})=\frac{p(l;\bm{\theta})}{p(S;\bm{\theta})}$ is a posterior distribution indicating the probability that a given set $S$ is generated by the path $l$. Therefore, $\nabla \log{p(S;\bm{\theta})}$ can be seen as the expected value of $\nabla p(l;\bm{\theta})$ where the random variable $l$ follows the posterior distribution $p(l \mid S;\bm{\theta})$.

We can apply Monte Carlo method to estimate $\nabla \log p(S;\bm{\theta})=E_{l\sim p(l|S;\bm{\theta})} [\nabla \log p(l;\bm{\theta})]$, where the key point is to sample generating paths from $p(l|S;\bm{\theta})$. A naive idea is to use rejection sampling, where we use the model with parameters $\bm{\theta}$ to generate random paths and reject those that cannot induce $S$. Obviously such a naive implementation would be extremely inefficient as the rejection rate would be very high. 

To sample $l$ from $p(l|S;\bm{\theta})$ more efficiently, we devise an important sampling method as follows. Let $\tau(a,s,S)$ be an indicator function, where $\tau(a,s,S)=1$ if $\exists l=[s_0, a_0, \dots, s_{T-1}, a_{T-1}, s_T]$ such that $l$ induces $S$ and $\exists i \in \{0,1,\dots,T-1\}$, $s_i=s \wedge a_i=a$. In our sampling algorithm, for a state $s_t$, we only consider actions $a_t$ such that $\tau(a_t,s,S) = 1$. We take such an action $a_t$ with probability 
$$
q(a_t \mid s; \bm{\theta},S)=\frac{p(a_t \mid s;\bm{\theta})}{\sum_{a:\tau(a,s,S) = 1}p(a \mid s;\bm{\theta})}.
$$ 
Therefore, the proposal distribution $q(l;\bm{\theta},S)$ for $l=[s_0, a_0, \dots, s_{T-1}, a_{T-1}, s_T]$ in our importance sampling is

\begin{equation*}\small
q(l;\bm{\theta},S) = q(a_0 \mid s_0;\bm{\theta},S)\times q(a_1 \mid s_2;\bm{\theta},S)\times \cdots \times q(a_{T-1} \mid s_{T-1};\bm{\theta},S)
\end{equation*}
The (unnormalized) importance weight of $l$ is 
\begin{equation*}\small
r(l;\bm{\theta}) = \left(\sum_{a:\tau(a,s_0,S)=1} p(a \mid s_0; \bm{\theta})\right) \times
\left(\sum_{a:\tau(a,s_1,S)=1} p(a \mid s_1; \bm{\theta})\right)\times
             \cdots
             \times \left(\sum_{a:\tau(a,s_{T-1},S)=1} p(a \mid s_{T-1};\bm{\theta})\right)
\end{equation*}
Clearly, we have
\begin{equation*}\small
r(l;\bm{\theta})q(l;\bm{\theta},S) \propto {p(a_0 | s_0;\bm{\theta})}{p(a_1 | s_1;\bm{\theta})}...{p(a_{T-1} | s_{T-1};\bm{\theta})} = p(l;\bm{\theta}).
\end{equation*}
This shows the correctness of the sampling method. 


\noindent \textbf{An EM perspective} Using our importance sampling algorithm to estimate $\nabla \log p(S;\bm{\theta})$ in minimizing the loss function $\mathcal{L}(\bm{\theta})$ can be interpreted from an Expectation-Maximization algorithm's perspective. Our method is actually a Monte Carlo EM algorithm if we treat a generating path $l$ as a latent variable.
In the E-step, we sample the latent variable $l$ from the posterior distribution $p(l|S;\bm{\theta}^{old})$. Our importance sampling can boost the efficiency of the E-step. In the M-step, we reconstruct an object function which is $\sum_i^M r(l_i)R^{-1}\log p(l_i;\bm{\theta}^{old})$, where $M$ is the number of samples and $R = \sum_{i=1}^M r(l_i)$. To maximize this function, we update $\bm{\theta}$ by gradient based methods, such as $\bm{\theta}^{new} = \bm{\theta}^{old} + \eta \sum_{i=1}^M r(l_i)R^{-1}\nabla_{\bm{\theta}} \log  p(l_i;\bm{\theta}^{old})$, where $\eta$ is the learning rate and $\sum_{i=1}^M r(l_i)R^{-1}\nabla_{\bm{\theta}} \log  p(l_i;\bm{\theta}^{old})$ is estimated by our importance sampling algorithm.

\noindent \textbf{MRW as an instance of Sequence2Set}
We show that the Markov Random Walk (MRW) model \citep{Tongwen21} is also an instance of our Sequence2Set model. 
The sequence model in MRW is a random walk on the item graph~\citep{Tongwen21} where a node is an item and we have a special stop item $\kappa$. The state $s_t$ at time $t$ is associated with a set $S_{s_t}$ which includes all the items visited until time $t$. An action $a_t$ taken at time $t$ given the state $s_t$ is to jump to an item $i_t$ from $i_{t-1}$, the item visited at time $t-1$. Once the stop item $\kappa$ is visited, the random walk terminates. The probability of jumping from $i_{t-1}$ to $i_t=i$ is parameterized as $p(i_t=i \mid s_{t-1}; \bm{\theta})$, where $\bm{\theta}$ is the embeddings of items. Specifically, in MRW, $p(i_t=i \mid s_{t-1}; \bm{\theta})=\frac{\exp (\bm{e}_{i}^\top \bm{e}_{i_{t-1}})}{\sum_j \exp (\bm{e}_{j}^\top \bm{e}_{i_{t-1}})}$, where $\bm{e}_i$ is the embedding vector of item $i$. When an action $a$ is jumping to an item $i \neq \kappa$, the indicator $\tau(a,s,S)=1$ if $i \in S$. When an action $a$ is jumping to the stop item $\kappa$, $\tau(a,s,S)=1$ if $S_s=S$. \cite{Tongwen21} propose an algorithm of complexity exponential to $|S|$ to calculate the gradient $\nabla \log{p(S;\bm{\theta})}$ for learning the item embeddings and thus, only small sets of size at most 4 are considered in \citep{Tongwen21}. If we adopt our important sampling algorithm, we can deal with large sets.

\section{GRU2Set and SetNN}\label{sec:models}
In this section, we propose two deep Sequence-to-Set models GRU2Set and Set Neural Networks (SetNN). 
We first introduce some general settings that will be used in both GRU2Set and SetNN.

\noindent \textbf{Item embedding and stop item}
We also adopt item embeddings, which should be learned from the training dataset, to parameterize the probability $p(a \mid s;\bm{\theta})$ in the sequence generative model. We assign an embedding vector $\bm{e}_i \in \mathbb{R}^d$ to each item $i$. Similar to MRW~\citep{Tongwen21}, we also add a stop item $\kappa$ where once we add $\kappa$ to the set associated to the current state, we stop the generating process. The stop item $\kappa$ also has an embedding vector $\bm{e}_{\kappa} \in \mathbb{R}^d$. 


\begin{wrapfigure}{r}{0.25\textwidth}
    \centering
  	\includegraphics[width=0.25\textwidth]{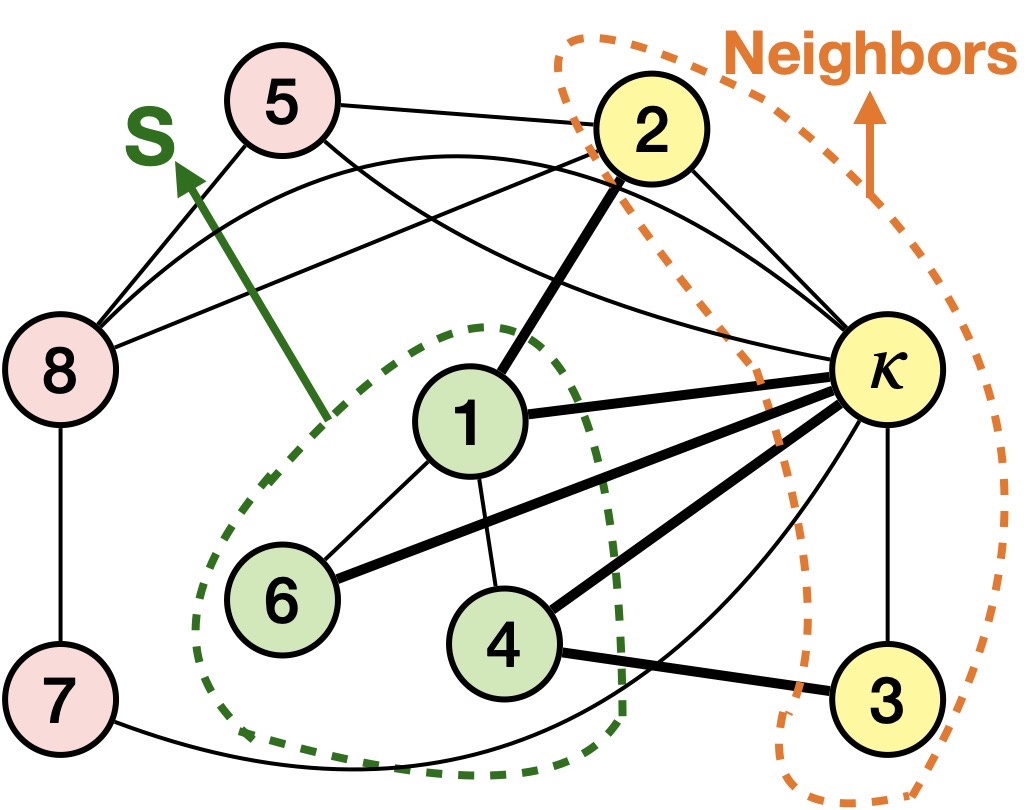}
    \caption{An example of sparse item graph. The item $\kappa$ is the stop item.}
    \label{fig:ItemGraph}
\end{wrapfigure}

\noindent \textbf{Sparse item graph} In our Sequence-to-Set model, the indicator function $\tau(a,s,S)$ is a key for us to efficiently sample generating paths from the posterior distribution $p(l \mid S; \bm{\theta})$. Intuitively, $\tau(a,s,S)$ indicates what actions are valid given the current state $s$ and the induced set $S$. To further boost the efficiency of our importance sampling algorithm, we make extra constraints on the items that can be added in the next step when we are given the current state $s$. To do so, we build an item graph $G=\langle V,E \rangle$, where $V$ is the set of items plus the stop item $\kappa$ and $(i,j) \in E$ if there is a set $S$ in the training dataset $\mathcal{S}$ such that $i \in S$ and $j \in S$. The stop item is connected to all the other items. When using our importance sampling algorithm, given the current state $s$ associated with a set $S_s \subseteq S$, $\tau(a,s,S)=1$ if action $a$ is adding an item $i \in S \cap \text{Neighbor}(S_s)$, where $\text{Neighbor}(S_s)$ is the set of neighbors to items in $S_s$. When starting from the first state $s_0$ associated with a set $S_{s_0}=\emptyset$, we set $\text{Neighbor}(S_{s_0})=V \setminus \{\kappa\}$. In real set datasets such as TMALL and HKTVMALL used in our experiments, such an item graph built from training datasets is often very sparse. Therefore, we call the item graph the {\bf sparse item graph}. Clearly, the sparsity of the item graph can help boost the efficiency of generating random sets from a learned model.


\noindent \textbf{Choice vector}
In our model design, the conditional probability $p(a \mid s; \bm{\theta})$ is the key to control the generating process. Exploiting the idea of representation learning, we represent each state $s$ with an embedding vector $\bm{c}_s \in \mathbb{R}^d$. We call $\bm{c}_s$ the {\bf choice vector} of $s$. Given $s$, we control the probability of adding $i$ as the next item as
\begin{equation*}\small
    p(i \mid s; \bm{\theta})=\frac{\exp (\bm{c}_s^\top \bm{e}_i)}{\sum_{j\in \text{Neighbor}(S_s)} \exp (\bm{c}_s^\top \bm{e}_j)}.
\end{equation*}
The sparsity of the item graph can make computing the normalization term in $p(i \mid s; \bm{\theta})$ efficient. Note that in MRW~\citep{Tongwen21}, the choice vector $\bm{c}_s$ is the embedding of the last item visited so far. We will see that in GRU2Set and SetNN, we use deep neural nets to aggregate the embeddings of all the items visited so far to obtain the choice vector $\bm{c}_s$. The initial choice vector $\bm{c}_{s_0}$ is trainable.



\subsection{GRU2Set}
We leverage GRU~\citep{Kyunghyun14} to build our first sequence-to-set model, since GRU is one of the most popular sequential model based on RNN and its computational cost is low. We set the history vector $H$ of GRU as a $d$-dimensional vector and $H$ is regarded as the choice vector of the current state. The initial history vector $H_0$ is trainable and $H$ is updated according to the design of GRU. The state-associated set $S_{s_t}$ for the state $s_t$ at time $t$ is just the set of items added in the first $t$ steps of the sequence generating process of GRU. When the stop item is added, we terminate the generating process. Due to limited space, the overview of GRU2Set is put in Appendix. 


\begin{figure}[h]
	\centering
  	\includegraphics[width=0.7\textwidth]{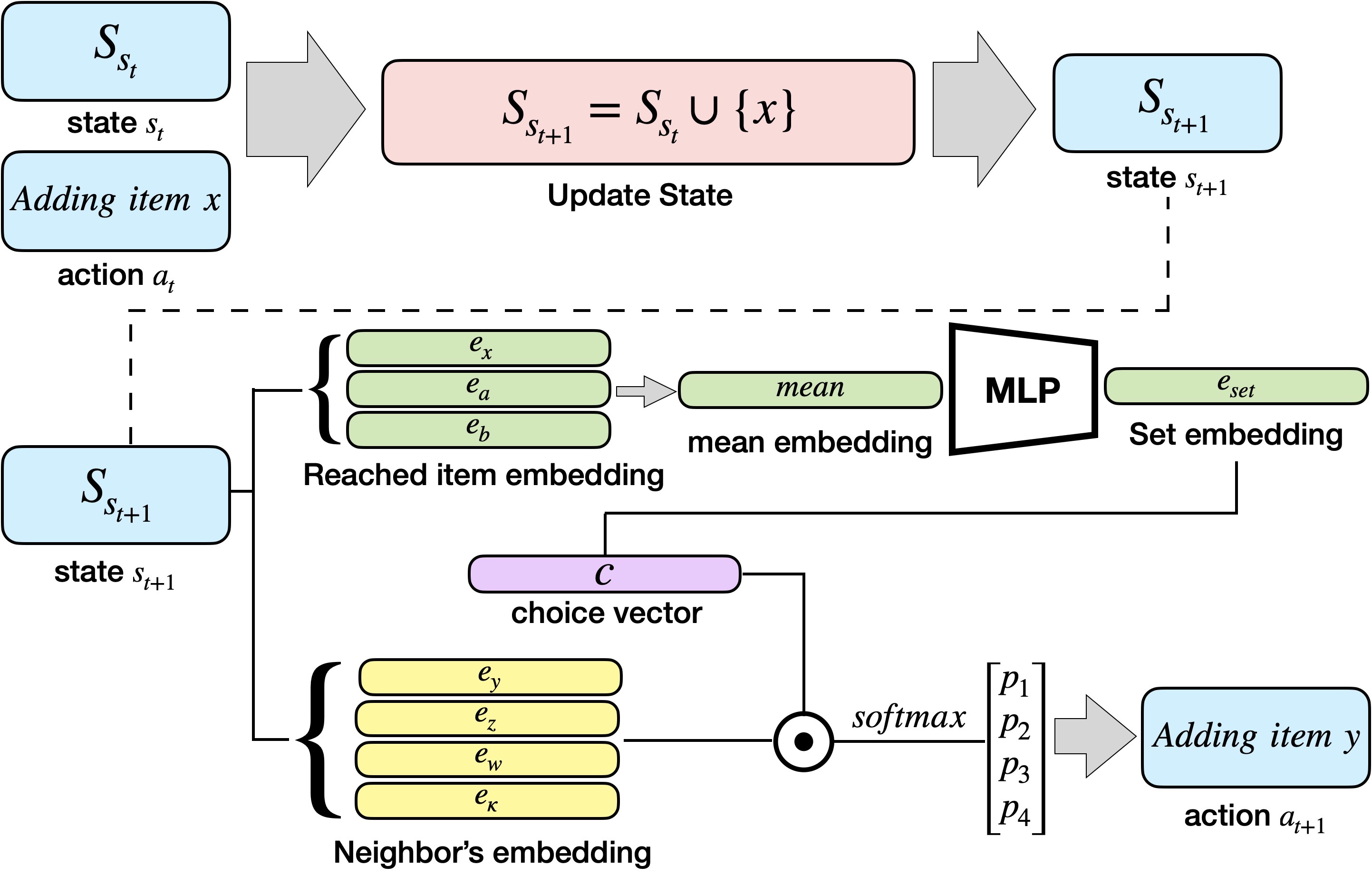}
    \caption{Overview of SetNN. We do not show the situation when we add the stop item. If the action is adding the stop item, we end the process immediately.}
    \label{fig:SetNN}
\end{figure}

\subsection{SetNN}
The choice vector $\bm{c}_s$ can be regarded as the embedding of the set $S_s$. In GRU2Set, the choice vector, which is the history vector $H$, is sensitive to the order of items added to $S_s$. As sets are permutation invariant, we hope to have a sequence-to-set model which can also produce permutation invariant representations of sets. Therefore, we develop the Set Neural Networks (SetNN) where the key idea is to design a permutation invariant aggregation of embeddings of items in a set. The following theorem by \cite{Zaheer17} guides our model design.


\begin{theorem}\label{theorem:permutation}
 (\citep{Zaheer17}) A function $f(X)$ operating on a set $X$ having elements from a countable universe, is a set function, i.e., invariant to the permutation of instances in $X$, iff it can be decomposed in the form $\rho(\sum_{x \in X} \Phi(x) )$  for suitable transformations $\Phi$ and $\rho$.
\end{theorem}

$\Phi(x)$ can be viewed as the embedding of $x$ and the sum over $\Phi(x)$ can be replaced by other aggregate operators such as mean or pooling. In SetNN, we use an MLP to express the function $\rho$ in Theorem~\ref{theorem:permutation} to obtain the embedding (choice vector) $\bm{c}_s$ of $S_s$ for any state $s$. Fig~\ref{fig:SetNN} shows the overview of SetNN.

\subsection{Expressive Power of GRU2Set and SetNN}\label{subsec:exp}

Compared to SetNN, our GRU2Set model uses the history vector $H$. By the expressive power of RNN \citep{Maximilian06}, as we increase the embedding dimension $d$, hidden neural units and layers in GRU module, the history vector $H$ of GRU2Set can approximate any function on a finite history path arbitrarily close. Then, of course, $H$ can express any permutation invariant function on sets which implies that the expressive power of GRU2Set is more than SetNN.

To explore the expressive power of SetNN, we first give a recursive definition of the probability that a set $S$ is generated by SetNN, denoted by $p(S)$, as follows.
\begin{equation}\label{equation:set probability SetNN}\small
\begin{aligned}
p(S) &= p(\kappa|S)\gamma(S)~~~~~\text{($\kappa$ is the stop item)}  \\
\gamma(S) &= \sum_{x \in S}p(x|S \setminus \{x\}) \gamma(S \setminus \{x\})\\
\gamma(\emptyset) &= 1,
\end{aligned}
\end{equation}
where $p(x|S \setminus \{x\})$ is the conditional probability of adding item $x$ to the set $S \setminus \{x\}$ and $\gamma(S)$ is the probability of reaching a state $s$ such that $S_s=S$ in SetNN. The following theorem shows that Eq.~\eqref{equation:set probability SetNN} is general enough to cover all possible set distributions.


\begin{theorem}\label{theorem:experss}
For any distribution $q(S)$ on all subsets of a ground set of items, there exists a group of transition probabilities $p(x|S)$ and $p(\kappa|S)$, such that $q(S) = p(S)$ holds for any $S$, where $p(S)$ is defined in Eq.~\eqref{equation:set probability SetNN}.
\end{theorem}

The proof of Theorem~\ref{theorem:experss} can be found in Appendix~\ref{section:proof}. As we increase the capacity of the MLP in SetNN, SetNN can represent any conditional probability $p(x|S)$ and $p(\kappa|S)$, and as a result of Theorem~\ref{theorem:experss}, SetNN and GRU2Set can express any set distributions.



\section{Size-Bias Trick for Improving Learned Set Distributions}\label{section:Size Distribution}

In this section, we propose a {\bf size-bias} trick that can help reduce the distance between the ground truth set distribution $p^*(S)$ and the learned distribution $p(S)$, where $p(S)$ can be learned by any generative model not limited to our Sequence-to-Set models. We first decompose the ground truth distribution $p^*(S)$ when $|S|=k$ as 
$$
    p^*(S)=p^{*(k)}(S)p^*_k,
$$
where $p^*_k=\sum_{A: |A|=k}p(A)$ is the probability of generating a set of size $k$, and $p^{*(k)}(S)=\frac{p(S)}{p^*_k}$ is the probability that a random size-$k$ set is $S$.
Suppose we know the real size distribution $p^*_k$ and we have $q^{(k)}(S)$ as an estimation of $p^{*(k)}(S)$ for each $k$. How can we construct a distribution $q(S)$ close to $p^*(S)$? An intuitive way is to combine $q^{(k)}(S)$ and $p^*_k$ such that $q(S) = q^{(k)}(S) p^*_k$ if $|S|=k$~\citep{stelzner20, Benson18}. However, we argue that this intuitive way is not always the optimal plan as follows.

Let $K$ be the largest possible set size. To find the best size distribution $q_k$ to collaborate with our estimation $q^{(k)}(S)$, we consider the following optimization aiming at minimizing the KL-divergence between the distribution $q(S) = q^{(k)}(S) q_k$ and the ground truth $p^*(S)$.



\begin{equation}\label{problem:best collaborator}\small
\begin{aligned}
\min_{q_1,\dots,q_K}~~& KL(q||p^*) = \sum_S q(S) \log \frac{q(S)}{p^*(S)}\\
s.t.\ \ & q(S) = q_k \times  q^{(k)}(S),\ \forall |S|=k\\
& 0 \le q_k \le 1, \forall k\\
& \sum_{k=1}^{K} q_k = 1\\
\end{aligned}
\end{equation}
Since we do not know $p^*(S)$, we cannot solve Eq.~\eqref{problem:best collaborator} exactly. However, we still can tell if setting $q_k = p^*_k$ is the optimal solution to some extend. 

Let $f_k(q_k) = \sum_{S:|S| = k} q(S) \log \frac{q(S)}{p^*(S)}$. We have $KL(q||p^*) = \sum_k f_k(q_k)$.
Note that $f_k(q_k)$ only depends on $q_k$. We can interpret Eq.~\eqref{problem:best collaborator} as a portfolio optimization problem. There are $K$ projects. $q_k$ is the investment for project $k$ and $- f_k(q_k)$ is the revenue of project $k$. To decide the best portfolio $q_k$, 
we check the derivative of $f_k(q_k)$, which indicates the marginal revenue. We have
\begin{equation*}\small
    f'_k(q_k) = 1 + \sum_{S:|S|=k}q^{(k)}(S) \log \frac{q(S)}{p^*(S)}
\end{equation*}
By taking $q_k = p^*_k$, we have
\begin{equation*}\small
f'_k(p^*_k) = 1 + \sum_{S:|S|=k}q^{(k)}(S) \log \frac{q^{(k)}(S)}{p^{*(k)}(S)}= 1 + KL\left(q^{(k)}(S) || p^{*(k)}(S)\right)\\
\end{equation*}
As the estimation quality of $q^{(k)}(S)$ often varies for different $k$, the derivative $f'_k(p^*_k)$ probably also varies for different $k$, making $q_k = p^*_k$ not a stationary point. Therefore, we probably need to set $q_k$ different from $p^*_k$ for solving Eq.~\eqref{problem:best collaborator}. We first make the following reasonable assumption.



\begin{assumption}\label{assumption:smaller k are easier}
Small-sized sets are easier to learn than large sets. In other words, $q^{(k)}(S)$ estimates $p^{*(k)}(S)$ better as the set size $k$ decreases.
\end{assumption}
Based on Assumption~\ref{assumption:smaller k are easier}, compared to $p^*_k$, we should ``invest'' more on small set sizes to construct the ``portfolio'' (size distribution) $q_k$. Therefore, we set $q_k$ slightly bigger than $p^*_k$ for small $k$, and set $q_k$ smaller than $p^*_k$ for big $k$. To address the issue of not knowing $p^*_k$, we can resort to the empirical size distribution of the training dataset, which is a good estimation of the real size distribution $p^*_k$. We construct a biased size distribution $q_k$ by adopting the following heuristic.

\noindent \textbf{Constructing Biased Size Distribution} Suppose the empirical size distribution of the training data $\mathcal{S}$ is $[p_1, p_2, ...,p_K]$. We first calculate the rest proportion by $r_k = p_k / (p_{k} + p_{k+1} + ... + p_K)$, which means $p_i = (1-r_1)...(1-r_{i-1})r_i$. Then we calculate $r'_k = \max\{ r_k + k \sqrt{|V|/ |\mathcal{S}|}, 1\}$, where $V$ is the set of all items. We construct a biased size distribution $[q_1,q_2,\dots,q_K]$ by setting $q_i =  (1-r'_1)...(1-r'_{i-1})r'_i$.

\section{Experiments}\label{sec:exp}
\paragraph{Datasets}
We did our empirical study on two real-world datasets of customer orders from online e-commerce platforms. This first dataset is TMALL (\url{https://www.tmall.com}) which contains \textbf{1363} items and orders from August 2018 to March 2019. We treated the collection of orders in each month as an instance and in total we have 8 instances for the TMALL dataset.
The second dataset is HKTVMALL (\url{https://www.hktvmall.com}) which contains 
\textbf{728} items form the supermarket sector and orders from February 2020 to September 2020. Similar to TMALL, we also treated HKTVMALL as 8 instances where each instance is the collection of orders in each month. For both datasets, We split all the orders in a month to a training dataset $\mathcal{S}_{train}$ and a testing dataset $\mathcal{S}_{test}$, where the size of $\mathcal{S}_{train}$ is 100,000 for Tmall and 200,000 for HKTVmall. The datasets can be found in the source code. We report the sparsity of item graphs and statistics of the datasets in Table~\ref{table:item graph}. The set size distribution of each dataset is shown in Table~\ref{table:size distribution}.

\begin{table}\small
  \centering
  \caption{Basic statistics}
  \label{table:item graph}
  \begin{tabular}{ccccc}
    \toprule
    dataset  & \#item & \#training sets & average \#testing sets & average \#edge\\
    \midrule
    TMALL & 1,363 & 100,000 & 393,226 & 60,956\\
    HKTVMALL   & 782 & 200,000 & 616,652  & 53,853\\
  \bottomrule
\end{tabular}
\end{table}

\begin{table}\small
  \centering
  \caption{Size distribution of TMALL and HKTVMALL}
  \label{table:size distribution}
  \begin{tabular}{cccccccc}
    \toprule
      & 1  & 2 & 3 & 4 & 5 & 6 & $\ge 7$ \\
    \midrule
    TMALL  & 38\%  & 18\% & 14\% & 10\% & 7\% & 5\% & 8\% \\
    HKTVMALL  & 42\%  & 17\% & 11\% & 8\% & 6\% & 5\% & 12\% \\
  \bottomrule
\end{tabular}
\end{table}

\paragraph{Experiment Setting and Evaluation}
We treat the method of directly using the histogram of $\mathcal{S}_{train}$ as the \textbf{benchmark} method, as it is a natural and unbiased method to estimate the set distribution. Note that this benchmark method has a major disadvantage that its cannot model probabilities of any sets not showing in $\mathcal{S}_{train}$.
We also compare our models with two baselines which are DCM (Discrete Choice Model)~\citep{Benson18} and MRW~\citep{Tongwen21}.

Following~\citep{Tongwen21}, for each method (except the benchmark), we generated a collection $\mathcal{S}_{pred}$ of 10,000,000 random sets and used the empirical set distribution of $\mathcal{S}_{pred}$ to approximate the set distribution learned by the model. 
We regarded the empirical distribution of the testing data $\mathcal{S}_{test}$ as the pseudo ground truth. Since the bigger $\mathcal{S}_{test}$ is, the more accurate the pseudo ground truth is, we set the size of $\mathcal{S}_{test}$ bigger than $\mathcal{S}_{train}$ as shown in Table~\ref{table:item graph}.  
Denote by $N_{test}(S)$ the number of sets that are $S$ in $\mathcal{S}_{test}$. Let $N_{pred}(S)$ be the number of sets that are $S$ in $\mathcal{S}_{pred}$. We used the $\ell_1$-distance between a model's (approximate) distribution and the pseudo ground truth to evaluate the effectiveness of the model. Specifically, the $\ell_1$-distance can be calculated as follows.
\begin{equation}\small
    \ell_1(\mathcal{S}_{test}, \mathcal{S}_{pred}) = \sum_{S \in \mathcal{S}_{test} \cup \mathcal{S}_{pred} } |\frac{N_{test}(S)}{|\mathcal{S}_{test}|} - \frac{N_{pred}(S)}{|\mathcal{S}_{pred}|}|
\end{equation}
We set the embedding dimension of all methods as 10. For each training and testing dataset, the biased size distribution was the same for all methods and was obtained by the heuristic introduced at the end of Section~\ref{section:Size Distribution}. For all experiments, the MLP of SetNN only has one hidden layer which contains 50 neural units. All the experiments were ran on a CPU with 10 cores. The optimizer used by us is RMSProp with default parameter in PyTorch. The source code and data used in our experiments can be found at \url{https://github.com/LongtaoTang/SetLearning}.

\paragraph{Experiment Results}

\begin{table}[h]\small
  \centering
  \caption{The performance of each groups on TMALL. G0 stands for the first group of data.}
  \label{table:Tmall}
  \begin{tabular}{ccccccccccccc}
    \toprule
     & G0 & G1 &  G2 & G3 & G4 & G5 & G6 & G7 & Average & STDEV\\
    \midrule
    Benchmark   & 1.11 & 1.02 & 0.91 & 0.90 & 0.94 & 0.94 & 0.93 & 0.85 & 0.95 & 0.07\\
    +Size Bias & 1.06 & 0.97 & 0.87 & 0.86 & 0.90 & 0.90 & \textbf{0.89} & 0.81 & 0.91 & 0.07\\
    \midrule
    DCM        & 1.17 & 1.08 & 0.95 & 0.94 & 1.00 & 0.99 & 0.99 & 0.89 & 1.00 & 0.08\\
    +Size Bias & 1.10 & 1.01 & 0.90 & 0.89 & 0.94 & 0.93 & 0.93 & 0.84 & 0.94 & 0.07\\
    \midrule
    MRW        & 1.14 & 1.07 & 0.99 & 0.96 & 1.02 & 1.02 & 1.03 & 0.92 & 1.02 & 0.06\\
    +Size Bias & 1.06 & 0.97 & 0.88 & 0.87 & 0.90 & 0.92 & 0.95 & 0.84 & 0.92 & 0.06\\
    \midrule
    SetNN      & 1.12 & 1.01 & 0.92 & 0.89 & 0.90 & 0.91 & 1.01 & 0.89 & 0.96 & 0.08\\
    +Size Bias & \textbf{1.04} & \textbf{0.93} & \textbf{0.84} & \textbf{0.82} & \textbf{0.85} & \textbf{0.87} & \textbf{0.89} & \textbf{0.79} & \textbf{0.88} & 0.07\\
    \midrule
    GRU2Set    & 1.11 & 1.04 & 0.89 & 0.90 & 0.94 & 0.94 & 1.00 & 0.90 & 0.97 & 0.07\\
    +Size Bias & \textbf{1.02} & \textbf{0.93} & \textbf{0.83} & \textbf{0.82} & \textbf{0.85} & \textbf{0.86} & \textbf{0.89} & \textbf{0.80} & \textbf{0.87} & 0.06\\
  \bottomrule
\end{tabular}
\end{table}

\begin{table}[h]\small
  \centering
  \caption{The performance of each groups on HKTVMALL. G0 stands for the first group of data.}
  \label{table:HKTVmall}
  \begin{tabular}{ccccccccccc}
    \toprule
     & G0 & G1 &  G2 & G3 & G4 & G5 & G6 & G7 & Average & STDEV\\
    \midrule
    Benchmark  & 0.73 & 0.87 & 0.91 & 0.84 & 0.81 & 0.87 & 0.81 & 0.80 & 0.83 & 0.05\\
    +Size Bias & \textbf{0.70} & 0.83 & 0.88 & 0.81 & \textbf{0.79} & 0.84 & 0.78 & \textbf{0.77} & 0.80 & 0.05\\
    \midrule
    DCM        & 0.76 & 0.89 & 0.93 & 0.88 & 0.84 & 0.89 & 0.84 & 0.82 & 0.86 & 0.05\\
    +Size Bias & 0.72 & 0.84 & 0.89 & 0.83 & 0.80 & 0.85 & 0.79 & 0.78 & 0.81 & 0.05\\
    \midrule
    MRW        & 0.80 & 0.91 & 0.94 & 0.88 & 0.88 & 0.90 & 0.85 & 0.85 & 0.88 & 0.04\\
    +Size Bias & 0.74 & 0.84 & 0.88 & 0.82 & 0.83 & 0.85 & 0.79 & 0.80 & 0.82 & 0.04\\
    \midrule
    SetNN      & 0.82 & 0.86 & 1.00 & 0.84 & 0.89 & 0.92 & 0.84 & 0.84 & 0.88 & 0.05\\
    +Size Bias & \textbf{0.71} & \textbf{0.82} & \textbf{0.87} & \textbf{0.79} & \textbf{0.79} & \textbf{0.83} & \textbf{0.76} & 0.78 & \textbf{0.79} & 0.04\\
    \midrule
    GRU2Set    & 0.83 & 0.87 & 0.94 & 0.84 & 0.93 & 0.94 & 0.86 & 0.82 & 0.88 & 0.05\\
    +Size Bias & \textbf{0.70} & \textbf{0.80} & \textbf{0.85} & \textbf{0.78} & \textbf{0.79} & \textbf{0.82} & \textbf{0.75} & \textbf{0.75} & \textbf{0.78} & 0.04\\
  \bottomrule
\end{tabular}
\end{table}


Due to limited space, we only report the main experimental results. More detailed experimental results can be found in Appendix~\ref{sec:sg} and \ref{sec:size}.

Table~\ref{table:Tmall} and Table~\ref{table:HKTVmall} show the experimental results. We place the results of using the size-bias trick below the original results. We find that both DCM and MRW never beat the benchmark method, no matter the size-bias trick is played or not. A possible reason is that in our experiments we did not make size constraints on sets, while \cite{Tongwen21} constrained the set size to be at most 4 and \cite{Benson18} set the maximum size of sets to be 5. Big sets in our data may have negative effects on learning both DCM and MRW.

We can see that applying the size-bias trick can always reduce the $\ell_1$-distance significantly, which demonstrates the effectiveness of this trick. Before applying the size-bias trick, the benchmark method has the best performance which is slightly better than our models GRU2Set and SetNN. However, after applying the size-bias trick, our GRU2Set model becomes the best and it outperforms the benchmark method by 4\% on TMALL and 2.5\% on HKTVMALL. SetNN also outperforms the benchmark method after using the size-bias trick. This suggests that GRU2Set and SetNN may learn probabilities of small sets better than the benchmark method.

The SetNN model performs slightly worse than GRU2Set but the gap between them is small. A possible reason is that GRU2Set has stronger expressive power than SetNN as illustrated in Section~\ref{subsec:exp}. GRU2Set records the order of items added to the set, while SetNN ignores such order to achieve permutation invariant set embeddings. This shows that GRU2Set includes more information in the generating process than SetNN. However, we want to emphasize that SetNN can produce permutation invariant set embeddings while GRU2Set cannot. If we have a downstream task that needs set embeddings, SetNN may have more advantages in this case.




\section{Conclusion}
In this paper, we present a Sequence-to-Set method to build generative models for set data and the SOTA method MRW is an instance of our method. To utilize deep learning in learning set distributions, we further design two models, GRU2Set and SetNN, which are two instances of our Sequence-to-Set model. To learn better set distributions from data, we also propose a size-bias trick. Experimental results on two e-commerce order datasets clearly show that our models outperform the baselines including two SOTA methods in this line of research.

For future work, we will explore using more sophisticated deep learning modules in our model, such as replacing the MLP in SetNN with Set Transformer and substituting the RNN in GRU with more advanced sequence models. 
For potential negative societal impacts, learning the embeddings of items and sets might cause user privacy leakage as many other embeddings might do. Research on how to hide personal information on set data could help protect users' privacy.

\section*{Acknowledgments and Disclosure of Funding}
Tang and Yang's research is supported in part by the Hong Kong Research Grants Council under ECS grant 21214720, City University of Hong Kong under Project 9610465, and Alibaba Group through Alibaba Innovative Research (AIR) Program. Zhou's research is supported in part by City University of Hong Kong under Project 7200694. The authors thank the HKTVmall Open Databank for providing the HKTVMALL dataset. All opinions, findings, conclusions, and recommendations in this paper are those of the authors and do not necessarily reflect the views of the funding agencies.

\bibliographystyle{plainnat}
\bibliography{Reference}

\newpage
\appendix
\section{Overview of GRU2Set}
\begin{figure}[ht]
	\centering
  	\includegraphics[width=0.8\textwidth]{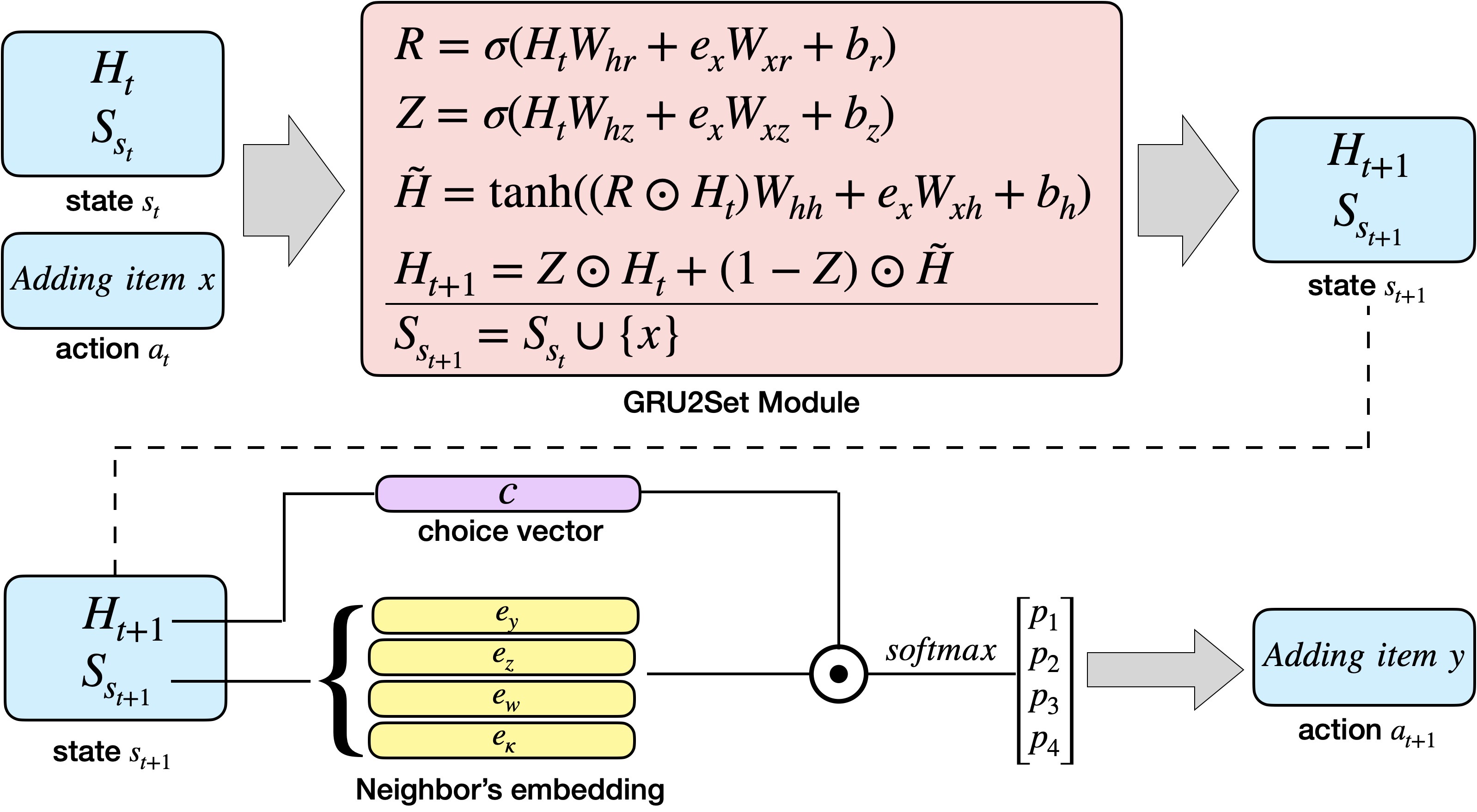}
    \caption{Overview of GRU2Set. We do not show the situation when we add the stop item. If the action is adding the stop item, we end the process immediately.}
    \label{fig:GRU2Set}
\end{figure}

\section{Proof of Theorem 2}\label{section:proof}
Without loss of generality, we assume that the probability of the empty set in $q(\cdot)$ is always 0. Let $V$ be the set of all items. We prove the theorem by an induction on $|V|$.

Basis Step: When $|V| = 1$, there is only one item in $V$ and suppose the item is $v_1$. Then $q(\{v_1\}) = 1$. We can set $p(v_1|\emptyset)=1, p(\kappa|\{v_1\}) =1$. Then $p(\{v_1\}) = p(v_1|\emptyset) \cdot p(\kappa|\{v_1\})=1$, which means $q=p$.

Inductive Step: Suppose the theorem holds for $|V| = m$. We discuss the case when $|V| = m + 1$.

Suppose $v_i$ is the $i$-th item. Let $V = V' \cup \{v_{m+1}\}$ and $V' = \{v_1, v_2, ..., v_m\}$. We define two sets $\mathcal S_{include} = \{S \big| v_{m+1} \in S\}$ and $\mathcal S_{exclude}= \{S \big| v_{m+1} \notin S\}$

First, we consider the conditional distribution $q_1$ on the condition that $v_{m+1} \notin S$. Then 
$$q_1(S) = \frac{q(S)}{\sum_{A \in \mathcal S_{exclude}} q(A)}$$

Note that $q_1$ is a distribution over all subsets of $V'$ and $|V'| = m$. From the inductive hypothesis, we can build a distribution $p_1$ by setting up $p_1(\kappa|S)$ and $p_1(v|S)$ for $v\in V'$ and $S \in S_{exclude}$, such that $p_1 = q_1$.

For $S\in \mathcal S_{exclude} \setminus \{\emptyset\}$ and $v \in V'$, we set 
$$p(\kappa|S) = p_1(\kappa|S)$$
$$p(v|S) = p_1(v|S)$$

and
$$p(v|\emptyset) = p_1(v|\emptyset) \sum_{A \in \mathcal S_{exclude}} q(A)$$

Now, we show that, for any $S\in \mathcal S_{exclude}$, we have $p(S) = q(S)$.
$$p(S) = p_1(S) \times \sum_{A \in \mathcal S_{exclude}} q(A) = q_1(S) \times \sum_{A \in \mathcal S_{exclude}} q(A) = q(S)$$

Second,  we consider $S \in \mathcal S_{include}$. We set 
$$p(v_{m+1}|\emptyset) = \sum_{A \in \mathcal S_{include}} q(A)$$
$$p(\kappa|\{v_{m+1}\}) = \frac{q(\{v_{m+1}\})}{\sum_{A \in \mathcal S_{include}} q(A)}$$

For $S = \{v_{m+1}\}$, we have
$$p(\{v_{m+1}\}) = p(v_{m+1}|\emptyset) \times p(\kappa|\{v_{m+1}\}) = q(\{v_{m+1}\})$$

For $S \in \mathcal S_{include} \setminus \{\{v_{m+1}\} \}$, we can map it to $S \setminus \{v_{m+1}\}$. We define

$$\mathcal S'_{include} = \left\{S \setminus \{v_{m+1}\} \big| S \in \mathcal S_{include} \setminus \{\{v_{m+1}\} \}\right\}$$

The conditional distribution $q_2$ on the condition $v_{m+1} \in S$ is a distribution over $\mathcal S'_{include} $. Then, for any $S \in \mathcal S'_{include}$,

$$q_2(S) = \frac{q(S \cup \{v_{m+1}\})}{\sum_{A \in \mathcal S'_{include}} q(A \cup \{v_{m+1}\})}$$

Note that $q_2$ is a distribution over all subsets of $V'$ and $|V'| = m$. From the inductive hypothesis, we can build a distribution $p_2$ by setting up $p_2(\kappa|S)$ and $p_2(v|S)$ for $v\in V'$ and $S \in S_{exclude}$, such that $p_2 = q_2$.

For $S\in \mathcal S_{include} \setminus \{\{v_{m+1}\} \}\} $, we set
$$p(\kappa|S) = p_2(\kappa|S \setminus \{v_{m+1}\} )$$
$$p(v|S) = p_2(v|S \setminus \{v_{m+1}\} )$$

and
$$p(v|\{v_{m+1}\}) = p_2(v|\emptyset) \times \sum_{A \in \mathcal S'_{include}} q(A \cup \{v_{m+1}\}).$$

For any $S\in \mathcal S_{include} \setminus \left\{\{v_{m+1}\} \right\} $, we have 
\begin{displaymath}
\begin{aligned}
p(S) & = p_2(S \setminus \{v_{m+1}\}) \times \sum_{A \in \mathcal S'_{include}} q(A \cup \{v_{m+1}\}) \\
     & = q_2(S \setminus \{v_{m+1}\}) \times \sum_{A \in \mathcal S'_{include}} q(A \cup \{v_{m+1}\}) \\
     & = q(S)\\
\end{aligned}
\end{displaymath}

Therefore, we have constructed $p(v|S)$ and $p(\kappa|S)$ such that $p(S)=q(S)$ for any $S \subseteq V$. The theorem holds when $|V| = m+1$. This completes the induction. 

\section{Optimal Size Distribution under $\ell_1$-Distance}\label{section:other distance}

In Section~\ref{section:Size Distribution}, we want to minimize the KL-divergence between $q(S)$ and $p^*(S)$. Here we discuss the situation when we try to minimize the $\ell_1$-distance. The constraints are the same as those in minimizing KL-divergence, while the objective becomes
\begin{equation*}
    \min_{q_1,\dots,q_K}  \ell_1(q, p^*) = \sum_S ||q(S) - p^*(S)||_1
\end{equation*}
Then we define
\begin{equation*}
    f_k(q_k) = \sum_{|S|=k} ||q(S) - p^*(S)||_1
\end{equation*}

Similar to the discussion in Section~\ref{section:Size Distribution}, we check the derivative $f_k'(p_k^*)$,
\begin{equation*}
    f_k'(q_k) = \sum_{|S|=k} q^{(k)}(S) \text{sgn}[q_k q^{(k)}(S) - p_k^* p^{(k)*}(S)]
\end{equation*}
Here $\text{sgn}[\cdot]$ is the sign function\footnote{$\text{sgn}[x]=1$ when $x > 0$; $\text{sgn}[x]=0$ when $x = 0$; $\text{sgn}[x]=-1$ when $x < 0$.}. By taking $q_k = p^*_k$, we have
\begin{equation*}\label{eq:dr_l1}
    f_k'(p^*_k) = \sum_{|S|=k} q^{(k)}(S) \text{sgn}[q^{(k)}(S) - p^{*(k)}(S)]
\end{equation*}
Therefore, probably $f_k'(p^*_k)$ varies for different $k$, if the distance between $q^{(k)}$ and $p^{*(k)}$ varies for different $k$. As a result, $q_k=p^*_k$ probably is not the optimal size distribution for minimizing the $\ell_1$-distance under Assumption 1. Moreover,
\begin{enumerate}
    \item When $q^{(k)}$ and $p^{*(k)}$ are far away form each other. Then when $q^{(k)}(S)$ is large, $p^{*(k)}(S)$ is small. It will bring a large positive term to $f'_k(p^*_k)$. If $q^{(k)}(S)$ is small and $p^{*(k)}(S)$ is large, it will bring a small negative term. Thus $f'_k(p^*_k)$ is large when $q^{(k)}$ and $p^{*(k)}$ is far away form each other.
    \item When $q^{(k)}$ and $p^{*(k)}$ are close, intuitively, $f_k'(p^*_k)$ should be close to 0.
\end{enumerate}

\begin{figure}[htbp]
    \centering
    \subfigure[Uniform]{
        \includegraphics[width=0.45\textwidth]{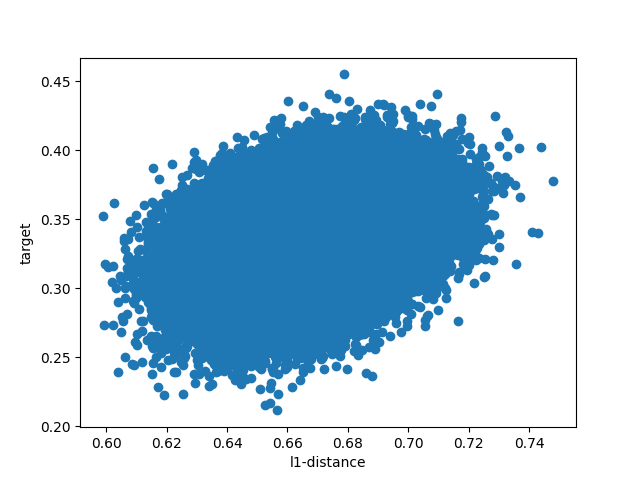}
    }
    \subfigure[Gaussian]{
        \includegraphics[width=0.45\textwidth]{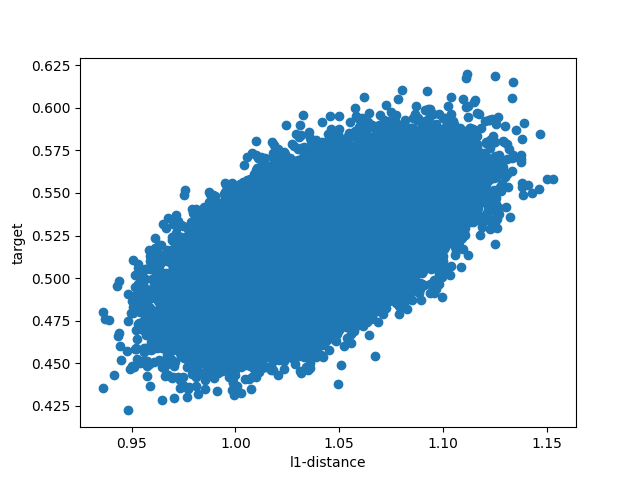}	
    }
    \caption{$\ell_1$-distance $\ell_1(q^{(k)}(S), p^{*(k)}(S))$ v.s. $\sum_{i=1}^{1000} q^{(k)}_i \text{sgn}[q^{(k)}_i - p^{*(k)}_i]$. ``target'' indicates the value of $\sum_{i=1}^{1000} q^{(k)}_i \text{sgn}[q^{(k)}_i - p^{*(k)}_i]$.}\label{Fig:pairs}
\end{figure}

Besides the above intuitive analysis, we also employ numerical simulations to explore the relationship between $f_k'(p^*_k) = \sum_{|S|=k} q^{(k)}(S) \text{sgn}[q^{(k)}(S) - p^{*(k)}(S)]$ and $\ell_1(q^{(k)}(S), p^{*(k)}(S))$. We randomly generate the two distributions $q^{(k)}$ and $p^{*(k)}$ as follows. 
\begin{enumerate}
    \item We treat $q^{(k)}=(q^{(k)}_1,q^{(k)}_2,\dots,q^{(k)}_{1000})$ and $p^{*(k)}=(p^{*(k)}_1,p^{*(k)}_2,\dots,p^{*(k)}_{1000})$ as two 1000-dimensional vectors. We randomly and independently generate each entry of $q^{(k)}$ and $p^{*(k)}$. 
    \item Entries of $q^{(k)}$ and $p^{*(k)}$ are both sampled from a uniform distribution $U[0,1]$. We also employ a standard Gaussian distribution $\mathcal{N}(0,1)$ to generate the random entries. After sampling all entries, we normalize both $q^{(k)}$ and $p^{*(k)}$ to make them probability distributions.
    \item We generate 100,000 pairs of random $q^{(k)}$ and $p^{*(k)}$. For each pair, we calculate $\ell_1(q^{(k)}(S), p^{*(k)}(S))$ and $\sum_{i=1}^{1000} q^{(k)}_i \text{sgn}[q^{(k)}_i - p^{*(k)}_i]$. Then we plot all the 100,000 pairs in Fig.~\ref{Fig:pairs}.
\end{enumerate}

Fig.~\ref{Fig:pairs} clearly shows the positive correlation between $\ell_1(q^{(k)}(S), p^{*(k)}(S))$ and value of Eq.~\eqref{eq:dr_l1} using $q^{(k)}$ and $p^{*(k)}$ as the inputs. 

The above discussion and simulation results demonstrate that if $q^{(k)}$ is a bad estimation of $p^{(k)}$, we need to invest less on $k$. This is similar to the case of minimizing the KL-divergence discussed in Section 5.

\section{Generalization Ability of Sparse Item Graph}\label{sec:sg}

One may question if the sparse item graph constructed from training samples is too restricted such that many possible sets of items cannot be generated by it. We report the ratios of testing sets that cannot be generated by the sparse item graph in Table~\ref{table:Sparse Tmall} and Table~\ref{table:Sparse HKTVmall}. It can be seen that on average only less than 2\% testing sets cannot be generated by the sparse item graph built upon training samples. The reason may be that the number of possible pairwise co-occurrences of items is much less than $O(n^2)$, since many pairs of items are rarely purchased together if the two items have very different functions. 

\begin{table}[h]\small
  \centering
  \caption{Ratio of testing sets that cannot be generated by the sparse item graph constructed from training data on TMALL.}
  \label{table:Sparse Tmall}
  \begin{tabular}{c|ccccc|c}
    \toprule
    \multirow{2}*{Group} & \multicolumn{5}{c|}{Order Size} & \multirow{2}*{Total Ratio}\\
    \cline{2-6}
      & 1 & 2 &  3 & 4 & $\ge$5 &  \\
    \midrule
    0 & 0.0000 & 0.0058 & 0.0036 & 0.0023 & 0.0038 & 0.0154 \\
    1 & 0.0000 & 0.0063 & 0.0036 & 0.0023 & 0.0038 & 0.0160 \\
    2 & 0.0000 & 0.0078 & 0.0043 & 0.0024 & 0.0034 & 0.0180 \\
    3 & 0.0000 & 0.0080 & 0.0045 & 0.0025 & 0.0041 & 0.0190 \\
    4 & 0.0000 & 0.0089 & 0.0047 & 0.0024 & 0.0038 & 0.0197 \\
    5 & 0.0000 & 0.0085 & 0.0046 & 0.0027 & 0.0040 & 0.0198 \\
    6 & 0.0000 & 0.0111 & 0.0059 & 0.0031 & 0.0039 & 0.0240 \\
    7 & 0.0000 & 0.0106 & 0.0056 & 0.0032 & 0.0041 & 0.0234 \\
    \midrule
    mean & 0.0000 & 0.0084 & 0.0046 & 0.0026 & 0.0039 & 0.0194 \\
  \bottomrule
\end{tabular}
\end{table}

\begin{table}[h]\small
  \centering
  \caption{Ratio of testing sets that cannot be generated by the sparse item graph constructed from training data on HKTVMALL.}
  \label{table:Sparse HKTVmall}
  \begin{tabular}{c|ccccc|c}
    \toprule
    \multirow{2}*{Group} & \multicolumn{5}{c|}{Order Size} & \multirow{2}*{Total Ratio}\\
    \cline{2-6}
     & 1 & 2 &  3 & 4 & $\ge$5 &  \\
    \midrule
    0 & 0.0000 & 0.0018 & 0.0005 & 0.0002 & 0.0002 & 0.0027 \\
    1 & 0.0000 & 0.0014 & 0.0004 & 0.0001 & 0.0002 & 0.0022 \\
    2 & 0.0000 & 0.0013 & 0.0004 & 0.0001 & 0.0002 & 0.0020 \\
    3 & 0.0000 & 0.0012 & 0.0003 & 0.0002 & 0.0002 & 0.0018 \\
    4 & 0.0000 & 0.0014 & 0.0004 & 0.0002 & 0.0002 & 0.0023 \\
    5 & 0.0000 & 0.0009 & 0.0003 & 0.0001 & 0.0002 & 0.0015 \\
    6 & 0.0000 & 0.0010 & 0.0003 & 0.0001 & 0.0002 & 0.0016 \\
    7 & 0.0000 & 0.0013 & 0.0004 & 0.0001 & 0.0002 & 0.0020 \\
    \midrule
    mean & 0.0000 & 0.0013 & 0.0004 & 0.0002 & 0.0002 & 0.0020 \\
  \bottomrule
\end{tabular}
\end{table}

\section{Size-wise Analysis}\label{sec:size}

We report more detailed experimental results in this section. The results can help readers better understand the advantages of our model and the effect of our size-bias trick. Moreover, based on the results, we will show that we can even combine Histogram with our models to build even stronger models for learning set distributions.

\begin{table}[h]\small
  \centering
  \caption{The average size distribution on TMALL}
  \label{table:Size distributione Tmall}
  \begin{tabular}{c|cccccc}
    \toprule
    \multirow{2}*{Method} & \multicolumn{5}{c}{Order Size} \\
    \cline{2-6}
     & 1 & 2 &  3 & 4 & $\ge$5\\
    \midrule
    Size-bais & 0.5015 & 0.2608 & 0.1577 & 0.0654 & 0.0146\\
    Benchmark & 0.3844 & 0.1780 & 0.1363 & 0.1033 & 0.1980\\
    GRU2Set & 0.3755 & 0.1916 & 0.1394 & 0.0999 & 0.1937\\
    SetNN & 0.3815 & 0.2097 & 0.1366 & 0.0913 & 0.1809\\
    MRW & 0.3525 & 0.1966 & 0.1392 & 0.0983 & 0.2134\\
    DCM & 0.3844 & 0.1780 & 0.1363 & 0.1033 & 0.1980\\
  \bottomrule
\end{tabular}
\end{table}

\begin{table}[h]\small
  \centering
  \caption{The average size distribution on HKTVMALL}
  \label{table:Size distributione HKTVmall}
  \begin{tabular}{c|cccccc}
    \toprule
    \multirow{2}*{Method} & \multicolumn{5}{c}{Order Size} \\
    \cline{2-6}
     & 1 & 2 &  3 & 4 & $\ge$5\\
    \midrule
    Size-bais & 0.4947 & 0.2228 & 0.1424 & 0.0904 & 0.0498\\
    Benchmark & 0.4155 & 0.1680 & 0.1088 & 0.0819 & 0.2258\\
    GRU2Set & 0.4001 & 0.1702 & 0.1092 & 0.0813 & 0.2392\\
    SetNN & 0.4048 & 0.1800 & 0.1155 & 0.0814 & 0.2183\\
    MRW & 0.4004 & 0.1938 & 0.1321 & 0.0912 & 0.1826\\
    DCM & 0.4155 & 0.1680 & 0.1088 & 0.0819 & 0.2258\\
  \bottomrule
\end{tabular}
\end{table}

We first report the average size distribution of each method as well as the size distribution calibrated by our size-bias trick in Table~\ref{table:Size distributione Tmall} and Table~\ref{table:Size distributione HKTVmall}. We can see that before applying the size-bias trick, all methods learn similar size distributions which are very close to the size distribution of the training data (Histogram). The effect of applying our size-bias trick is to increase the ratio of small-sized orders and reduce the number of large-sized orders in our prediction $\mathcal{S}_{pred}$.

\begin{table}[h]\small
  \centering
  \caption{The average size-wise Overlap on TMALL}
  \label{table:Size-wise Tmall}
  \begin{tabular}{c|ccccc|c}
    \toprule
    \multirow{2}*{Method} & \multicolumn{5}{c|}{Order Size} & \multirow{2}*{Total Overlap}\\
    \cline{2-6}
     & 1 & 2 &  3 & 4 & $\ge$5 &   \\
    \midrule
    Benchmark & 0.3692 & 0.1166 & 0.0325 & 0.0056 & 0.0010 & 0.5249\\
    +Size Bias & 0.3797 & 0.1267 & 0.0337 & 0.0049 & 0.0004 & 0.5454\\
    \midrule
    GRU2Set & 0.3494 & 0.1185 & 0.0407 & 0.0076 & 0.0010 & 0.5172\\
    +Size Bias & 0.3764 & 0.1359 & 0.0449 & 0.0053 & 0.0002 & 0.5627\\
    \midrule
    SetNN & 0.3475 & 0.1247 & 0.0407 & 0.0070 & 0.0008 & 0.5207\\
    +Size Bias & 0.3751 & 0.1349 & 0.0448 & 0.0052 & 0.0002 & 0.5602\\
    \midrule
    MRW & 0.3334 & 0.1129 & 0.0366 & 0.0067 & 0.0008 & 0.4904\\
    +Size Bias & 0.3682 & 0.1253 & 0.0398 & 0.0046 & 0.0002 & 0.5381\\
    \midrule
    DCM & 0.3691 & 0.1033 & 0.0242 & 0.0025 & 0.0004 & 0.4995\\
    +Size Bias & 0.3797 & 0.1198 & 0.0266 & 0.0018 & 0.0001 & 0.528\\
    \midrule
    hybrid(Benchmark + GRU2Set) & 0.3797 & 0.1359 & 0.0449 & 0.0053 & 0.0002 & \textbf{0.5660}\\
    hybrid(Benchmark + SetNN) & 0.3797 & 0.1349 & 0.0448 & 0.0052 & 0.0002 & 0.5648\\
  \bottomrule
\end{tabular}
\end{table}

\begin{table}[h]\small
  \centering
  \caption{The average size-wise Overlap on HKTVMALL}
  \label{table:Size-wise HKTVmall}
  \begin{tabular}{c|ccccc|c}
    \toprule
    \multirow{2}*{Method} & \multicolumn{5}{c|}{Order Size} & \multirow{2}*{Total Overlap}\\
    \cline{2-6}
     & 1 & 2 &  3 & 4 & $\ge$5 &  \\
    \midrule
    Benchmark & 0.4080 & 0.1271 & 0.0365 & 0.0103 & 0.0029 & 0.5848\\
    +Size Bias & 0.4142 & 0.1349 & 0.0388 & 0.0106 & 0.0021 & 0.6006\\
    \midrule
    GRU2Set & 0.3788 & 0.1253 & 0.0409 & 0.0116 & 0.0032 & 0.5598\\
    +Size Bias & 0.4091 & 0.1388 & 0.0470 & 0.0125 & 0.0023 & 0.6097\\
    \midrule
    SetNN & 0.3792 & 0.1268 & 0.0414 & 0.0111 & 0.0030 & 0.5615\\
    +Size Bias & 0.4064 & 0.1370 & 0.0461 & 0.0119 & 0.0022 & 0.6036\\
    \midrule
    MRW & 0.3836 & 0.1248 & 0.0403 & 0.0107 & 0.0028 & 0.5622\\
    +Size Bias & 0.4041 & 0.1319 & 0.0419 & 0.0107 & 0.0020 & 0.5906\\
    \midrule
    DCM & 0.4079 & 0.1225 & 0.0316 & 0.0074 & 0.0021 & 0.5715\\
    +Size Bias & 0.4112 & 0.1283 & 0.0338 & 0.0075 & 0.0019 & 0.5827\\
     \midrule
    hybrid(Benchmark + GRU2Set) & 0.4142 & 0.1388 & 0.0470 & 0.0125 & 0.0023 & \textbf{0.6148}\\
    hybrid(Benchmark + SetNN) & 0.4142 & 0.1370 & 0.0461 & 0.0119 & 0.0022 & 0.6114\\
  \bottomrule
\end{tabular}
\end{table}

We then report the effectiveness of each method on orders of different sizes in Table~\ref{table:Size-wise Tmall} and Table~\ref{table:Size-wise HKTVmall}. To better understand the result, we use the metric $Overlap$~\citep{Tongwen21} which can be converted to $\ell_1$-distance since
\begin{equation*}
    Overlap(\mathcal{S}_{test}, \mathcal{S}_{pred}) = \sum_{S \in \mathcal{S}_{test} \cup \mathcal{S}_{pred} } \min\{ \frac{N_{test}(S)}{|\mathcal{S}_{test}|}, \frac{N_{pred}(S)}{|\mathcal{S}_{pred}|}\}=\left(1-\ell_1(\mathcal{S}_{test}, \mathcal{S}_{pred})\right)*2
\end{equation*}
We decompose $Overlap(\mathcal{S}_{test}, \mathcal{S}_{pred})$ as
$$
Overlap(\mathcal{S}_{test}, \mathcal{S}_{pred})=o_1+o_2+o_3+o_4+o_5,
$$
where $o_k=\sum_{|S|=k} \min\{\frac{N_{test}(S)}{|\mathcal{S}_{test}|}, \frac{N_{pred}(S)}{|\mathcal{S}_{pred}|}\}$ for $k=1,2,3,4$, and
$o_5 = \sum_{|S| \geq 5} \min\{\frac{N_{test}(S)}{|\mathcal{S}_{test}|}, \frac{N_{pred}(S)}{|\mathcal{S}_{pred}|}\}$.

Using the results in Table~\ref{table:Size-wise Tmall} and Table~\ref{table:Size-wise HKTVmall}, as well as the size distributions reported in Table~\ref{table:Size distributione Tmall} and Table~\ref{table:Size distributione HKTVmall}, we can calculate the ''cost-effectiveness'' of guessing size-$k$ sets in $\mathcal{S}_{pred}$, which is the ratio between the overlap on size-$k$ sets and the ratio of size-$k$ sets in $\mathcal{S}_{pred}$. By simple calculation we find that the cost-effectiveness of small sets is often greater than that of large sets. This is not surprising as in both TMALL and HKTVMALL small sets are more abundant than large sets. In addition, the number of possible small sets is much less than the number of possible large sets due to the combinatorial explosion of sets. As a result, it is easier to learn the probabilities of small sets than large sets, which is consistent to our Assumption~\ref{assumption:smaller k are easier}. Therefore, if we increase the number of small sets in $\mathcal{S}_{pred}$ and reduce the number of large sets in $\mathcal{S}_{pred}$, the marginal gain on the overlap on small sets has a good chance to be greater than our loss on the overlap on large sets. This motivates us to apply our size-bias trick.

Another important finding from Table~\ref{table:Size-wise Tmall} and Table~\ref{table:Size-wise HKTVmall} is that Histogram's $o_1$ is often the best while our models' $o_2$, $o_3$ and $o_4$ are better than those of Histogram. We explain this from a perspective of the effective sample size and the number of possible size-$k$ sets. 

Take TMALL as an example. In our training data, on average we have $100,000*38.44\%=38,440$ size-1 orders. Compared to the number of items (1,363), 38,440 training samples are enough for us to learn the set distribution for size-1 sets well. Moreover, the distribution of size-1 sets actually is a standard discrete distribution. \cite{kamath2015learning} proves that for learning a discrete distribution, Histogram can achieve the min-max rate and is the asymptotically optimal method. Therefore, Histogram often achieves the best overlap on size-1 sets.

However, when it comes to sets of multiple items, the number of training samples is not enough compared to the number of possible size-$k$ sets for $k>1$. Moreover, the distribution of sets of multiple items is more than a standard discrete distribution, as similar sets may have some connections to each other. Such information is totally ignored by Histogram but well captured by our models. Note that Histogram has no extrapolation ability as it cannot model the probability of any unseen set. Therefore, it is not surprising that our models have better overlap on size-2, size-3 and size-4 sets than Histogram.

The baseline methods MRW~\citep{Tongwen21} and DCM~\citep{Benson18} both try to catch the connections between similar sets but their performance is not good. A possible reason is that both MRW and DCM make too strong assumptions on how random sets are generated, while our models adopt deep neural nets whose expressive power is extremely strong.

\paragraph{Hybrid Model} One interesting finding based on the detailed experimental results is that we can even combine Histogram with our models to build stronger models for learning set distributions. Specifically, for TMALL and HKTVMALL, we use Hitogram to construct the probabilities of size-1 sets and use our models for sets of multiple items. We call such a model a Hybrid model and we report its performance in Table~\ref{table:Size-wise Tmall} and Table~\ref{table:Size-wise HKTVmall}. Since the hybrid model has the advantages of both Histogram and our models, it achieves the best performance in Table~\ref{table:Size-wise Tmall} and Table~\ref{table:Size-wise HKTVmall}.


\end{document}